\documentclass[11pt]{article}
\usepackage{eamt26}
\usepackage{times}
\usepackage{url}
\usepackage{multirow}
\usepackage{latexsym}
\usepackage{amsmath}
\usepackage{lipsum}
\usepackage[small,bf]{caption}
\usepackage{inconsolata}
\usepackage{pgfplots}
\usepackage{pgffor}
\usepgfplotslibrary{external}
\pgfdeclarelayer{background}
\pgfsetlayers{background,main}
\usepackage{booktabs}
\usepackage{graphicx}
\usepackage{listings}
\usepackage{tabularx}
\usepackage{array}
\pgfplotsset{compat=1.18}

\lstdefinestyle{promptstyle}{
  basicstyle=\ttfamily\footnotesize,
  columns=fullflexible,
  breaklines=true,
  frame=single,
  framesep=0.6em,
  framerule=0.4pt,
  xleftmargin=0.8em,
  xrightmargin=0.8em,
  showstringspaces=false,
  keepspaces=true,
  lineskip=0.2em,
  aboveskip=1em,
  belowskip=1em
}

\title{Emotion Profiling in LLM-Based Literary Translation: \\Systematic Shifts Across MT and Post-Editing}

\author{
  Antonio Castaldo$^{1,2}$, Johanna Monti$^{1}$, Sheila Castilho$^{3}$ \\
  $^{1}$University of Naples ``L'Orientale'' \quad
  $^{2}$University of Pisa \quad
  $^{3}$Dublin City University \\
  \texttt{antonio.castaldo@phd.unipi.it},
  \texttt{jmonti@unior.it},\\
  \texttt{sheila.castilho@adaptcentre.ie}
}

\date{}

\begin{document}
\maketitle

\begin{abstract}
This paper investigates whether LLM translations exhibit identifiable emotional profiles and how post-editing reshapes them toward human-like norms. We compare LLM translations of Margaret Atwood's Oryx and Crake with their post-edited versions and a human translation, using a large-scale corpus of contemporary Italian science-fiction as a baseline. We examine emotion through lexicon-based and multilingual modeling, conducting a fine-grained analysis of emotional variation across systems. We find that MT systems introduce model-specific and statistically significant emotional fingerprints across translations, leading to a limited preservation of an author's voice.
\end{abstract}

\section{Introduction}

Large Language Models (LLMs) have quickly become the predominant approach to machine translation (MT), particularly in literary translation, where the scope extends beyond mere transfer of meaning, encompassing the preservation of an author's voice, style and their aesthetic intent \cite{zhang_how_2025}.

Recent research in the field of LLMs has demonstrated significant improvements in their ability to handle figurative language \cite{castaldo_prompting_2024,raunak_gpts_2023} and context-related issues \cite{castilho_online_2023}. These advancements make them especially promising for literary translation, where such features are fundamental. However, despite these improvements, the extent to which they may be able to preserve an author's voice remains underexplored. This gap becomes even more relevant when considering post-editing (PE). While it is generally assumed that PE brings MT outputs closer to HT in terms of quality, it remains unclear to what extent PE preserves the author’s voice present in the source text, or whether it is influenced by stylistic interference introduced by the MT system.

In this work, we study how authorial voice is rendered through the lens of emotional profiling, investigating whether different MT paradigms introduce distinct emotional patterns and how these are transformed during post-editing. The author's emotional voice, therefore, refers to the way the author intended emotions to be expressed in the original text. We conduct a comparative study on a set of translations of the literary novel ``Oryx and Crake'' by Canadian author Margaret Atwood \cite{atwood_oryx_2004}, from English to Italian. We analyze outputs from three LLM systems, alongside their post-edited versions and a published human translation (HT).

To complement this analysis and evaluate the robustness of our findings, we use a large-scale corpus of contemporary Italian science fiction from Wattpad. From this corpus, we sample 1M tokens and generate parallel translations using an LLM (TowerPlus-2B) and an NMT system (NLLB-200-600M). This large-scale setup allows us to move beyond individual examples and test for statistically significant patterns in emotional variation. Using lexicon-based and multilingual modeling approaches, we examine emotion shifts between source texts and translations, i.e. the differences in how emotionally charged language is rendered, finding that both systems introduce systematic and statistically significant changes in emotional properties.

Our findings highlight that MT systems exhibit distinctive emotional fingerprints, reflected in systematic differences in how emotional content is distributed, intensified, or attenuated throughout the text. The effectiveness of post-editing in restoring the author's voice appears to be strongly constrained by the quality of the underlying MT model. We further find these emotional fingerprints to be statistically significant and particularly concentrated at emotionally salient peaks, where the author's emotional voice would ideally be most faithfully preserved. 

\section{Related Work}

In the field of translation studies, a large amount of scientific literature has focused on whether translated texts exhibit features that are statistically different from naturally occurring human writing \cite{koppel_translationese_2011,volansky_features_2015,mauranen_translation_2004}. The research approach based on corpus linguistics is grounded in the pioneering work done by Baker~\shortcite{baker_corpus_1993}, who introduced the notion of ``translation universals''. According to the author, translation universals are hypotheses of linguistic features common to translated texts regardless of the source and target languages, a phenomenon often referred to as \textit{translationese}. These findings laid the methodological foundation for stylometric analysis of translated texts, leading to the hypothesis that if human translators leave identifiable fingerprints, MT systems should too potentially in more systematic ways. 

To this end, subsequent studies have confirmed the hypothesis, finding that MT systems display traces of translationese \cite{bizzoni_how_2020}. In relation to NMT, Vanmassenhove et al.~\shortcite{vanmassenhove_lost_2019} specifically documented a loss of lexical diversity in NMT outputs across several language pairs, potentially representing a universal fingerprint. This was confirmed by a later study, where the authors identified the tendency of NMT systems to exacerbate frequent patterns \cite{vanmassenhove_machine_2021}. Despite the growing body of literature in the field, the stylistic effects of post-editing on LLM outputs are less understood.

Studies on post-editing have typically focused on NMT systems where the concept of \textit{post-editese} has been introduced as a distinct register that characterizes post-edited translations, and differs from both MT outputs and human translations. Post-editing has been found to introduce intermediate stylometric profiles and stylistic normalisation \cite{toral_post-editese_2019}. More recent work, grounded in Baker's theory, revealed evidence of post-editese in NMT outputs of literary texts, finding that post-edited translations come closer to the MT outputs than to human translations in terms of stylistic features \cite{castilho_post-editese_2022}.

Most studies on translationese and post-editing have focused on surface-level features, such as lexical diversity and frequency distributions. While these metrics provide valuable insights, in the context of literary translation, they provide only a partial view of style. In literary texts, an author’s voice is not only reflected in structural and lexical choices, but also in the distribution and intensity of emotions conveyed throughout the narrative. In this regard, researchers have developed resources to identify and analyze emotions in text \cite{mohammad-turney-2010-emotions,passaro_2017}, in both monolingual and multilingual settings. Lexicon-based methods have been demonstrated to produce accuracies
comparable to machine learning and transformer- based methods, with the additional benefit of transparency and interpretability \cite{teodorescu_evaluating_2023,ohman_validity_2021}.

In this work, we make use of the available resources to conduct an emotion-based analysis of MT outputs. Specifically, we model the emotional profile of texts and investigate how it is transformed through the translation process. We extend this analysis to post-edited outputs, exploring whether post-editing mitigates or reinforces these emotional shifts, and whether post-editese exhibits distinctive emotional fingerprints.

\section{Experimental Setup}

We conduct two main experiments: one is a controlled study focused on a literary excerpt from Margaret Atwood's Oryx and Crake, of which we possess a human professional translation, three translations generated by different LLMs and three post-edited MT outputs, as described in Section~\ref{subsec:data-collection}. On this corpus, we rely on the Italian Emotive Lexicon \cite{passaro_2017} to compare the emotional density and intensity across the translation versions. Through this experiment, we aim to investigate whether MT systems introduce model-specific behavior in regards to the preservation, amplification or reduction of emotional voice in the text, as well as how post-editing affects these aspects.

The second study is a large-scale analysis on a corpus of naturally occurring literary texts, that we scrape from a fan-fiction platform called Wattpad. From Wattpad, we scrape one thousand stories consisting of 25M tokens in total. The stories were extracted from the relevant science-fiction category and have all been published in the last five years. For this second experiment, we adopt a more scalable approach using the NRC Emotion Lexicon \cite{mohammad-turney-2010-emotions} in its multilingual version.

\subsection{Data Collection}
\label{subsec:data-collection}
The translations analyzed in this study were obtained from a previous study, in which 121 segments of Margaret Atwood's Oryx and Crake were translated using different MT systems and subsequently post-edited by professional literary translators \cite{castaldo-etal-2025-extending}. We selected Oryx and Crake because the novel contains stylistically rich passages, figurative language, and neologisms, making it particularly suitable for studying emotional variation. Such characteristics allow for different renderings and creative deviations across translation versions. The diverse composition of the corpus, with professional translations and post-edited versions of different state-of-the-art MT systems, makes it an ideal resource for our study. We refer to that work for full details on the translation and post-editing pipeline. 

To empirically validate our emotion analysis, we compiled a large-scale reference corpus of contemporary Italian science-fiction. The corpus was constructed by scraping Wattpad, a major user-generated fiction platform, using the Python library Scrapy\footnote{\url{https://scrapy.org}}. We collected every available chapter from 1,000 stories belonging to the science-fiction genre, as identified by the platform's category system. Language identification was performed using langdetect\footnote{\url{https://pypi.org/project/langdetect/}}, retaining only texts classified as \textbf{Italian}, resulting in a corpus of approximately 25 million tokens. 

The Wattpad corpus allows us to assess whether the emotional profiles observed in MT, LLM, PE, and HT outputs reflect systematic tendencies of the models. By comparing these outputs against a broad sample of naturally occurring Italian science-fiction prose, we can more reliably identify which emotional shifts are consistently introduced, amplified, or suppressed by the different translation processes.

\subsection{Lexical Resources}

As mentioned in the previous section, two main lexical resources are used in this study: the Italian Emotive Lexicon \cite{passaro_2017} and the NRC Emotion Lexicon \cite{mohammad-turney-2010-emotions}. Both resources adopt the same eight basic emotions, modeled after Plutchik~\cite{plutchik_emotion_1985}, but diverge in the way they are composed and in the data they expose.

\paragraph{Italian Emotive Lexicon. } The Italian Emotive Lexicon was built from data collected from native Italian speakers via an online survey, where participants were asked to produce five words associated with each emotion. This list of seed words was then used to populate the lexicon. This lexicon provides a list of words connected to each of the eight emotions, with part-of-speech tagging and frequency data. Interestingly, it also exposes emotion centroids and the cosine similarity of each lexical item to the centroid.

\paragraph{NRC Emotion Lexicon. } The NRC Emotion Lexicon, on the other hand, was built for English through crowdsourcing where annotators labeled predefined words with emotions. The NRC Emotion Lexicon exposes a list of words with binary associations to each emotion. In addition to emotions, each word is also labeled for polarity (positive, negative). In the following versions, it has also been adapted to multilingual tasks, supporting over 108 languages, including Italian.

\section{Results}

In this section, we investigate whether different translation versions (HT, LLM, and PE) exhibit distinct emotional profiles, and to what extent these profiles converge toward HT. We focus on two main dimensions: the amount of emotion-related tokens present in the text compared to content words, and how strongly associated are the words with the emotion categories in the lexicon.

To this end, we first examine the number of tokens matched in the emotion lexicon relative to the total number of words. This provides an indication of how frequently emotion lexical items are used.

\begin{table}[h]
\centering
\small
\setlength{\tabcolsep}{5pt}
\begin{tabular}{l r r l r r}
\toprule
\multicolumn{3}{c}{\textbf{MT}} & \multicolumn{3}{c}{\textbf{PE}} \\
\cmidrule(r){1-3} \cmidrule(l){4-6}
\textbf{System} & \textbf{Match} & \textbf{Words} & \textbf{System} & \textbf{Match} & \textbf{Words} \\
\midrule
GPT-3.5   & 718 & 2175 & GPT-3.5   & 745 & 2279 \\
GPT-4     & 717 & 2190 & GPT-4     & 728 & 2232 \\
Mistral& 636 & 2026 & Mistral& 731 & 2253 \\
\midrule
\multicolumn{3}{l}{HT: 738 / 2218} \\
\bottomrule
\end{tabular}
\caption{Emotion-related matched tokens from the lexicon and word counts, per translation version.}
\label{tab:emotion-tokens}
\end{table}

Table~\ref{tab:emotion-tokens} provides an overview of the raw counts of emotion-related tokens across systems. If MT systems preserve the emotional profile of the source text, we would expect their values to be aligned with the human translation. On the opposite, systematic deviations may indicate model-specific tendencies.

As shown in Table~\ref{tab:emotion-tokens}, MT outputs differ in their degree of lexical alignment with the emotion lexicon. In particular, some systems produce fewer matched tokens, suggesting a tendency to under-represent emotion-related vocabulary, while others approach the levels observed in HT. Post-edited versions generally increase the number of matched tokens, indicating that human intervention tends to reintroduce or reinforce emotional lexical choices in line with HT. The lack of preservation of emotional lexicon could be attributable to accuracy errors in the translation or hallucinated outputs. We verify the alignment between the different MT versions, using translation quality metrics\footnote{The COMET model we use throughout our analysis is wmt-22-comet-da \cite{rei_comet-22_2022}.} and the HT as reference. We provide quality metrics results in Table~\ref{tab:auto-metrics}, divided per system and translation version.

\begin{table}[htbp]
\centering
\small
\setlength{\tabcolsep}{12pt}
\begin{tabular}{l r r r}
\toprule
\textbf{System} & \textbf{BLEU} & \textbf{chrF} & \textbf{COMET} \\
\midrule
\multicolumn{4}{c}{\textbf{MT}} \\
\cmidrule(lr){1-4}
GPT-4            & 28.63 & 53.55 & 0.79 \\
GPT-3.5          & 27.11 & 54.20 & 0.82 \\
Mistral          & 15.86 & 43.24 & 0.77 \\
\midrule
\multicolumn{4}{c}{\textbf{PE}} \\
\cmidrule(lr){1-4}
GPT-4       & 25.70 & 52.63 & 0.78 \\
GPT-3.5     & 25.29 & 53.05 & 0.78 \\
Mistral     & 21.24 & 49.02 & 0.79 \\
\bottomrule
\end{tabular}
\caption{Automatic evaluation scores across MT outputs and their post-edited (PE) versions.}
\label{tab:auto-metrics}
\end{table}

We observe from Table~\ref{tab:auto-metrics} that overall MT systems achieve comparable quality across all metrics, and even the systems receiving high scores exhibit deviations in terms of emotion lexicon from HT. The only observable exception is Mistral which is the worst performing system and also displays most divergence in terms of matched tokens, as displayed in Table~\ref{tab:emotion-tokens}.

\subsection{Emotion Analysis}
\label{subsec:emotion_analysis}

To provide a more fine-grained investigation of how emotion is distributed and how strongly it is expressed, we complement our analysis with calculations on emotional density and intensity. While the two metrics are certainly related, they measure different phenomena. \textit{Density} captures the proportion of emotion-lexicon matches among content words, whereas \textit{intensity} captures the average strength of those matches based on the cosine scores provided by the lexicon. Therefore, a sentence may have few words strongly associated with emotions, resulting in high intensity and low density, or many weakly associated words resulting in the opposite direction.

\subsection{Emotional Density}

We calculate \textit{emotional density} as the proportion of content words in a sentence with matched tokens from the lexicon. This measure serves as a proxy for how much emotion-related vocabulary is present in the sentence, independently of the specific emotion category involved. Emotional density shows HT to have the highest density, with the PE output from GPT-4 approaching it. While the PE outputs from Mistral and GPT-3.5 also show an increase in the number of matched tokens, their density increase is diluted by the higher number of content words. By aggregating the results per group (HT, MT, PE), PE improves the density of MT, making it approach HT. The post-edited version of GPT-4 is the translation that most approaches HT in terms of content words, emotion-related tokens and density. This may be attributed to the high performance of the LLM which in previous studies has proved to approach HT in terms of both quality \cite{hendy_how_2023} and creativity \cite{castaldo-etal-2025-extending}.

\begin{figure*}[h]
\centering
\includegraphics[width=\linewidth]{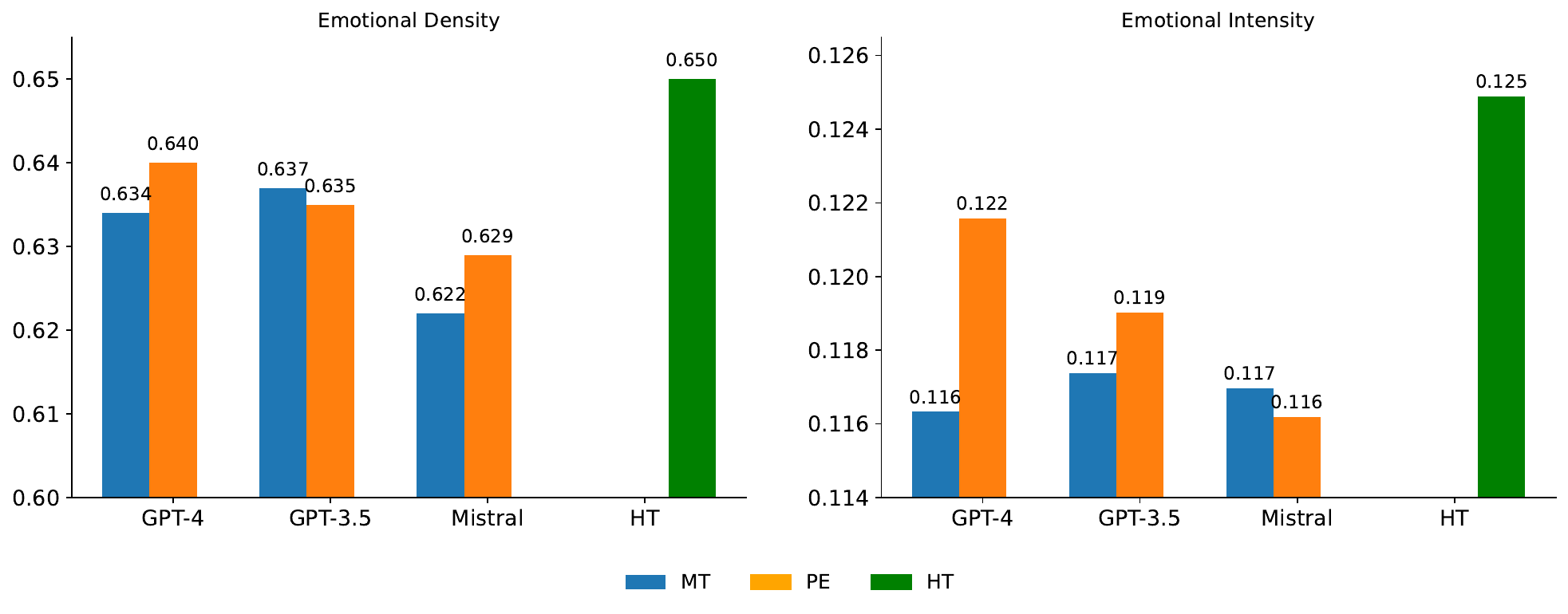}
\caption{Emotional density and intensity across MT systems, their post-edited (PE) versions, and human translation (HT).}
\label{fig:density-intensity}
\end{figure*}

\subsection{Emotional Intensity} 
\label{subsec:emotional-intensity}
\textit{Emotional intensity} measures how strongly the lexical choices in a sentence are associated with emotional content overall. To compute this score, first we match each lemmatized content word in the sentence to the lexicon, and retrieve its cosine similarity scores for the eight basic emotions. Then, we average these scores across all matched words for each emotion, obtaining sentence intensity as the mean of the resulting eight emotion scores. We show the results of both metrics side-by-side in Figure~\ref{fig:density-intensity} to allow easier comparison.

The results of emotional intensity agree with emotional density in terms of overall tendency. PE seems to increase the emotional density of the corpora, making the translations approach HT. Interestingly, we find a drastic increase in intensity after GPT-4 has been post-edited, suggesting that given the higher quality of the model's outputs it may be more easily adapted by the translators and lead to better overall preservation of the original author's voice. While the other models display only a marginal increase in intensity, the only exception is Mistral which exhibits a contraction in terms of intensity. Given that Mistral is the smallest and worst performing model according to quality metrics, the finding is consistent with our preliminary assumption that model performance affects the ability of the translators to adapt its content to maximize the preservation of emotions, as per the source.

\subsection{Emotion Preservation}

\begin{figure*}[h]
\centering
\includegraphics[width=\linewidth]{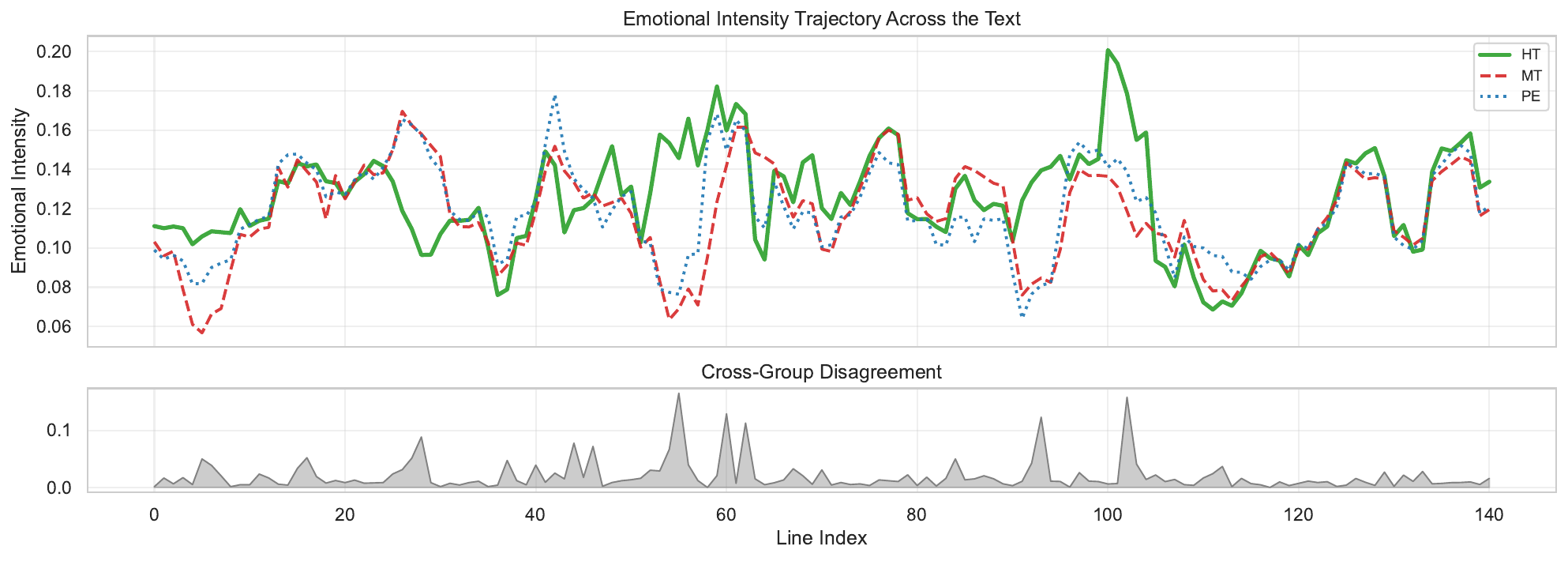}
\caption{Emotional trajectory across MT systems, their post-edited (PE) versions, and human translation (HT). Trajectory is smoothed using a 5-line rolling mean. The bottom of the figure shows per-sentence standard deviation, indicating disagreements across versions.}
\label{fig:intensity-trajectory}
\end{figure*}

By computing sentence-level emotional intensity over the corpora, we are able to trace an intensity trajectory across the text. To do so, we apply a 5-line rolling mean, then aggregate the resulting values per translation version. The per-sentence standard deviation, shown at the bottom of the figure, captures the degree of disagreement between versions at each point in the text. We present the resulting trajectories of HT, PE and MT in Figure~\ref{fig:intensity-trajectory}, where the horizontal axis represents the line index of the literary excerpt and the vertical axis the emotional intensity, allowing us to directly compare where the versions track closely and where they diverge the most.

Interestingly, we find that the majority of the disagreements is found in high-intensity peaks, which leads us to investigate how each model behaves in this regard. Therefore, to further investigate the relationship between model quality and emotional fidelity, we identify the twenty sentences with the highest emotional intensity in HT and measure how often each version preserves them. We define a peak as preserved when the translation intensity falls within one standard deviation of HT. Table~\ref{tab:peak-preservation} reports peak preservation rates across MT and PE systems out of the twenty identifed peaks.

Across MT outputs, we find preservation rates to be low, ranging from 38\% to 46\%, with no substantial difference between models. Post-editing consistently improves preservation, but the effect appears highly dependent on the model and does not mirror the trends observed for overall intensity. The post-edited GPT-4 achieves a modest gain over its MT output, but is outperformed by GPT-3.5 which reaches 64\%. We argue that this behavior could reflect a tendency in translation behavior induced by the MT quality. GPT-4's higher fluency and coherence may lead translators to introduce several small local adjustments across the text, that increase the overall intensity of the translation, while not directly addressing high-intensity peaks. This would be consistent with Castaldo et al.~\shortcite{castaldo-etal-2025-extending} that showed in their experiments how GPT-4 received more edits in post-editing, that required much fewer editing time. By contrast, GPT-3.5 may lead the translators to introduce more substantial revisions, recovering most of the emotional peaks found in HT. 

A further noteworthy result comes from the post-edited version of Mistral, which not only fails to improve over its MT output but drops to 31\%, the lowest rate across all systems. 

This finding reinforces the earlier assumption that low-quality outputs may hinder translators’ ability to recover emotional intensity, as their attention must first be directed toward resolving issues of accuracy and fluency before they can effectively address stylistic aspects.

\begin{table}[h]
\centering
\caption{Preservation of emotional intensity peaks across MT and PE versions. A peak is considered preserved if the translation intensity falls within one standard deviation of HT.}
\label{tab:peak-preservation}
\begin{tabular}{llcc}
\toprule
\textbf{System} & \textbf{Type} & \textbf{Preserved} & \textbf{Rate} \\
\midrule
GPT-3.5 & MT & 5/20 & 38\% \\
GPT-4   & MT & 6/20 & 46\% \\
Mistral & MT & 6/20 & 46\% \\
\midrule
GPT-3.5 & PE & 8/20 & 62\% \\
GPT-4   & PE & 7/20 & 54\% \\
Mistral & PE & 4/20 & 31\% \\
\bottomrule
\end{tabular}
\end{table}

\section{Large-Scale Emotion Analysis}
\label{sec:large-scale}

To assess whether the emotional patterns observed in Section~\ref{subsec:emotion_analysis} generalize beyond a single literary text, we conduct a large-scale validation experiment on a separate corpus of contemporary Italian science fiction drawn from Wattpad. We sample 1M tokens from this corpus and generate parallel translations from Italian into English using two systems: TowerPlus-2B \cite{alves_tower_2024}, an LLM-based MT system, and NLLB-200-600M \cite{nllbteam2022languageleftbehindscaling}, a neural MT system. Emotion detection is performed using the NRC multilingual lexicon (cf. §3.2) and then confronted with a valence-arousal-dominance (VAD) regression model, and a multilingual emotion classifier\footnote{\url{https://huggingface.co/Lsthf/multilingual-emotion-classification}} to ensure that the observed differences are not driven by the specific choice of resource. Emotion shift is calculated as the difference between the target and source emotion scores, normalized by text length.


Both systems introduce statistically significant shifts in emotion-related properties relative to the source ($p < 0.05$), with a consistent tendency to overproduce emotion-related tokens. TowerPlus generally amplifies emotion more than NLLB, though the direction of the shift depends on the representational framework used. At the level of emotion categories, TowerPlus systematically overproduces positive emotions relative to NLLB, with statistically significant differences for trust ($p = 0.0013$), surprise ($p = 0.0032$), and anticipation ($p = 0.0436$). These results confirm that different MT paradigms do not preserve emotion-related aspects of the source in the same way: while the magnitude of the differences is modest, they are systematic enough to be reliably detected across a corpus of this scale. This is consistent with the model-specific emotional profiles identified in the Oryx and Crake analysis, and suggests that the tendency of LLM-based systems to consistently introduce shifts in emotional content is not an artifact of a single text, but a more general property of the system.

The large-scale data further allows us to examine whether model differences are distributed across the text or concentrated at specific peaks, mirroring the pattern observed in Section~\ref{subsec:emotional-intensity} where the majority of disagreement was found in high-intensity peaks. Our results show that TowerPlus and NLLB tend to converge on low-emotion passages while diverging substantially in high-emotion source segments. 

\section{Discussion}

Taken together, the results reveal a set of consistent patterns. Overall, HT is the upper bound and PE consistently moves MT closer to HT, in terms of both emotional density and intensity, and in line with previous research on post-editing \cite{castilho_post-editese_2022}. However, the degree to which PE recovers emotional properties is not uniform across models, and the results suggest that MT output quality affects the translator's ability to intervene effectively on emotional content.

GPT-4 post-editing produces the largest increase in overall emotional intensity, with translators not only reintroducing more emotion-related vocabulary but selecting words more strongly associated with emotional content. We interpret this as evidence that higher-quality outputs are more easily adapted in terms of stylistic and emotional content, as translators may attend to more nuanced interventions rather than focusing on correcting fluency and accuracy errors.

While post-editing increases intensity and density across all models, this behavior does not always translate to better preservation of emotional peaks, as we see in the difference between the post-edited versions of GPT-4 and GPT-3.5 (cf. §4.4). GPT-4 is outperformed in peak preservation by GPT-3.5, and we argue that this divergence reflects a difference in how translators engage with the two outputs. The higher fluency of GPT-4 may encourage a more evenly distributed revision that raises the overall emotional density and intensity, without specifically addressing high-intensity peaks. GPT-3.5, on the other hand, may lead translators to introduce more substantial revisions, recovering emotional peaks in the process. This interpretation is consistent with Castaldo et al.~\shortcite{castaldo-etal-2025-extending}, who found that GPT-4 outputs attracted more numerous but less time-consuming edits, suggesting a pattern of small refinements across the translation, rather than substantial revisions.

This finding is consistent with the results obtained for Mistral. Although post-editing increases the model's emotional density, it results in a contraction of emotional intensity, and a peak preservation rate equal to 31\%, which is the lowest rate across all systems. This suggests that beyond a certain threshold of quality, translators may not be able to recover emotional content, while being confronted with more pervasive fluency and accuracy errors.

The large-scale analysis conducted on the Wattpad corpus situates these findings empirically, finding that the emotion shifts introduced by both TowerPlus and NLLB across 1M tokens are statistically significant and not specific to the document taken into analysis. The results confirm that the emotional shifts introduced by model-specific behavior are a systematic property of MT systems, leading to a partial preservation of an author's voice. Finally, our analysis confirms that model differences concentrate in high-emotion segments, suggesting that the emotional fingerprint of the model may be actually found in segments where an author's voice may need to be more faithfully preserved.

\section{Conclusion}

In this study, we investigated how emotions are distributed and expressed across different translation versions, starting from a fine-grained analysis on a literary excerpt from Margaret Atwood's Oryx and Crake and validating our findings with a large-scale analysis on a corpus of contemporary Italian science fiction drawn from Wattpad.

Our results consistently show that MT systems introduce model-specific emotional fingerprints that are detectable both at the level of individual translations and across large corpora. In the controlled study, we found that LLM and NMT outputs differ from HT in terms of emotional density and intensity, and that these differences are not uniformly distributed across the text but concentrate at emotionally salient peaks, where the preservation of an author's voice is most critical.

Post-editing consistently moves MT outputs closer to HT, but its effectiveness depends on the quality of the underlying system. Higher-quality outputs appear to facilitate stylistic adaptation leading to a higher preservation of emotional content, whereas lower-quality outputs redirect the translator's attention toward the correction of errors, leaving less room for emotional recovery. The large-scale analysis confirms our findings through an independent study, showing that the tendency of MT models to introduce model-specific emotional fingerprints, especially in high-intensity peaks, may be a systematic property of these models.

Taken together, these findings have direct implications for literary translations, suggesting that MT systems have distinctive emotional profiles and post-editing alone may not be sufficient to fully recover the emotional patterns observed in HT, especially when the underlying MT quality is low.

\section{Limitations}

Although we complemented the analysis with both neural and lexicon-based methods, the approaches adopted in this study remain approximations of emotional expression and may not fully capture its complexity.

We also acknowledge that our analysis focuses exclusively on translation versions without directly modeling the source text itself. As a result, our investigation of authorial voice is necessarily constrained by how that voice is rendered in translation, rather than by a direct comparison with the emotional properties of the original text.

In relation to the large-scale experiment, we cannot fully exclude the possibility that some of the stories extracted from Wattpad had already undergone machine translation or machine-assisted editing prior to publication on the platform. Although this risk is difficult to eliminate at scale, it may introduce noise into the comparison between naturally occurring texts and MT outputs.

Finally, while our findings suggest that MT systems exhibit distinctive emotional fingerprints, we do not evaluate reader perception. The observed emotional shifts are computationally measurable, but it remains unclear to what extent they are perceptible to readers or whether they significantly affect literary reception. Future work could therefore integrate reception studies and human evaluations to assess emotional fidelity in literary translation.

\section*{Acknowledgments}
This work has been funded by the Italian National PhD programme in Artificial Intelligence, partnered by University of Pisa and University of Naples ``L'Orientale'', through a doctoral grant (ID 39-411-24-DOT23A27WJ-6603) established by Ex DM 318, of type 4.1, co-financed by the National Recovery and Resilience Plan. 

The third author benefits from being member of the ADAPT SFI Research Centre at Dublin City University, funded by the Science Foundation Ireland under Grant Agreement No. 13/RC/2106\_P2.

\bibliography{custom,references}
\bibliographystyle{eamt26}

\end{document}